\pdfoutput=1

\documentclass[11pt]{article}

\usepackage{acl}

\usepackage{times}
\usepackage{latexsym}

\usepackage[T1]{fontenc}

\usepackage[utf8]{inputenc}

\usepackage{microtype}

\usepackage{inconsolata}
\usepackage{amsmath}
\usepackage{amsfonts}
\usepackage{graphicx}
\usepackage{subcaption}
\usepackage{enumitem}
\usepackage{booktabs}
\usepackage{caption}

\title{CA*: Addressing Evaluation Pitfalls in \\ Computation-Aware Latency for Simultaneous Speech Translation}

\author{
     Xi Xu\textsuperscript{\dag}, Wenda Xu\textsuperscript{\ddag}, Siqi Ouyang\textsuperscript{\dag}, Lei Li\textsuperscript{\dag} \\
    \textsuperscript{\dag}Carnegie Mellon University
    \textsuperscript{\ddag}University of California, Santa Barbara \\
    \texttt{\{xixu, siqiouya, leili\}@cs.cmu.edu} \\
    \texttt{wendaxu@ucsb.edu}
}

\begin{document}
\maketitle

\begin{abstract}

Simultaneous speech translation (SimulST) systems must balance translation quality with response time, making latency measurement crucial for evaluating their real-world performance. However, there has been a longstanding belief that current metrics yield unrealistically high latency measurements in unsegmented streaming settings. In this paper, we investigate this phenomenon, revealing its root cause in a fundamental misconception underlying existing latency evaluation approaches. We demonstrate that this issue affects not only streaming but also segment-level latency evaluation across different metrics. Furthermore, we propose a modification to correctly measure computation-aware latency for SimulST systems, addressing the limitations present in existing metrics.

\end{abstract}
\section{Introduction}
Simultaneous speech-to-text translation (SimulST) ~\cite{ma-etal-2020-simulmt} focuses on a real-time, low-latency scenario where the model starts generating the textual translation before the entire audio input is processed. Achieving high-quality translations with minimal latency is the primary objective of SimulST systems, with time constraints varying by scenario\cite{anastasopoulos-etal-2022-findings, agrawal-etal-2023-findings, ahmad-etal-2024-findings}. These constraints are typically quantified as latency, often defined as the average time delay between when a word is spoken and when its translation is generated. Accurate latency measurement is thus critical for evaluating system performance.

Various metrics have been introduced to measure SimulST system's latency, including Average Proportion (AP) \cite{cho2016neuralmachinetranslationsimultaneous}, Average Lagging (AL) \cite{ma-etal-2019-stacl}, Differentiable Average Lagging (DAL) \cite{cherry2019thinkingslowlatencyevaluation}, and Length Adaptive Average Lagging (LAAL)\cite{papi-etal-2022-generation} and Average Token Delay (ATD)\cite{kano2022average}. Recent studies \cite{papi-etal-2023-attention, papi2023alignatt, ahmad-etal-2024-findings, xu-etal-2024-cmus} have emphasized computation-aware latency as a more realistic way to evaluate SimulST performance in real-time scenarios where computation time cannot be ignored. 


As the performance of SimulST systems improves, researchers are motivated to apply them to unsegmented streaming long speech\cite{polak-2023-long, ouyang2024fasstfastllmbasedsimultaneous, papi-etal-2024-streamatt, iranzo-sanchez-etal-2021-stream-level}, which better represents real-world scenarios such as interpreting and lecture transcription. However, computation-aware latency scores reported by existing metrics have been observed to be unrealistically high, hindering progress in this area.

\begin{figure}[t]
    \centering
    \includegraphics[width=1\linewidth]{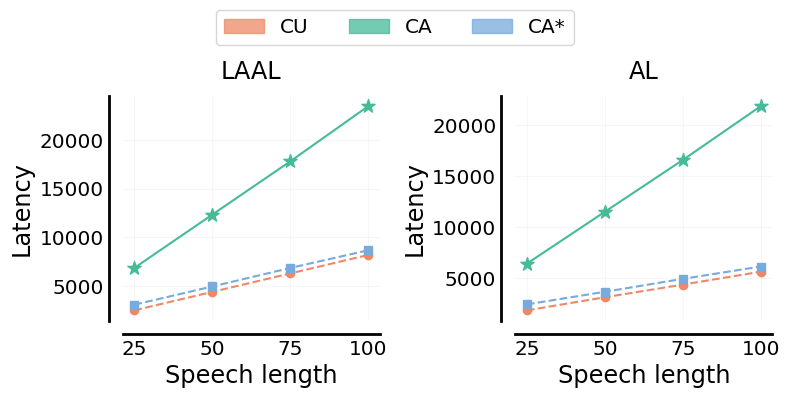}
    \caption{Computation-aware latency metrics tend to produce unrealistically high scores as speech duration increases. We segmented the original tst-COMMON dataset into 25s speech segments, then duplicated and concatenated them to create 50s, 75s, and 100s speech durations. The system used for evaluation is a wait-k-stride-n model, where n=3 and k=4, with each speech segment spanning 250 ms.}
    \label{fig:intro}
\end{figure}

As illustrated in Figure~\ref{fig:intro}, the computation-aware AL and LAAL metrics show drastic increases as the length of streaming speech increases. Specifically, the computation-aware LAAL increases from 6850 ms when the speech length is 25 seconds to 23460 ms when the speech length increased to 100 seconds. In contrast, the theoretical computation unaware LAAL remains mostly consistent, at 8181 ms. 


This paper aims to uncover the root causes of the unrealistic computation aware latency measurements and to present our proposed solution for correcting them. Section~\ref{sec:section2} describes the current latency metrics in detail, while Section ~\ref{sec:section3} outlines the misconceptions underlying these inconsistencies. Our proposed solution is introduced in Section
\ref{sec:section4}, where we address the limitations of existing evaluation methods and provide a refined approach to computation-aware latency measurement. 
In Section \ref{sec:section5}, we exemplify our proposed calculation aligned with a real-time setting, and show that this miscalculation also impacts pre-segmented speech.
\section{Latency Metrics}
\label{sec:section2}

\begin{figure}[t]
    \centering
\includegraphics[width=1\linewidth]{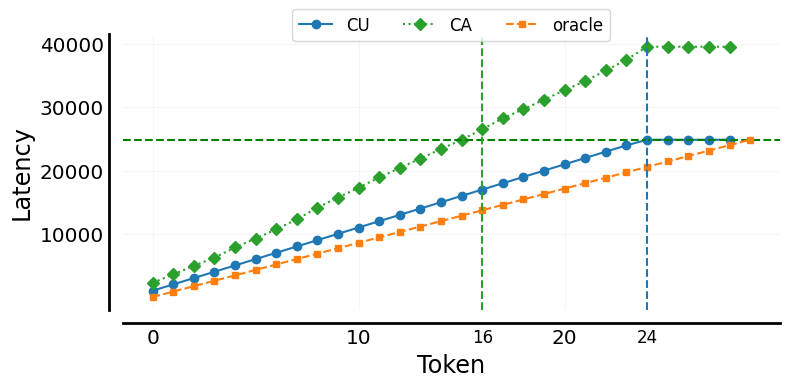}
\caption{Computation unaware and aware $d_i$ with the corresponding oracle delay $d^*$, where the intersection represents $\tau'(|\mathbf{X}|)$. (For illustration purposes, we plot only one token for three tokens in the stride-3 SimulST system.) After conversion, the latency AL\_CA only considers the first 46 outputs against the oracle, resulting from the unreliable calculation of computation elapsed, while AL\_CU considers all outputs until the last speech segment.}
\label{fig:fig2}
\end{figure}

An evaluation corpus for a speech translation task contains one or more instances, each consisting of a source speech sequence $\mathbf{X} = [x_1, ..., x_{|\mathbf{X}|}]$ and a reference text sequence $\mathbf{Y}^* = [y^*_1, ..., y^*_{|\mathbf{Y}^*|}]$. The system to be evaluated takes $\mathbf{X}$ as input and generates $\mathbf{Y} = [y_1, ..., y_{|\mathbf{Y}|}]$ as the target language's text translation incrementally. 

\begin{figure}[t]
    \centering
    \includegraphics[width=1\linewidth]{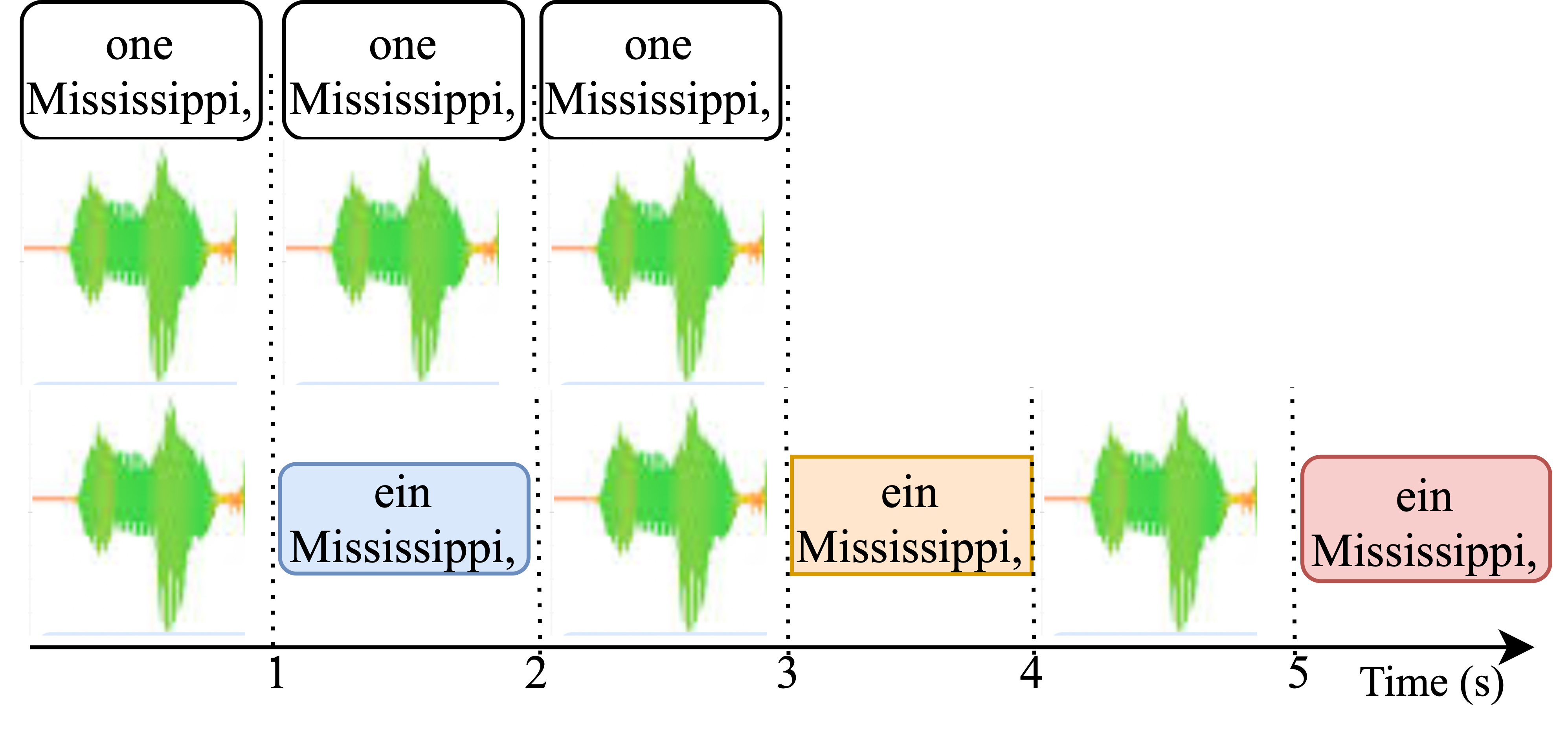}
    \caption{In practice, a SimulST system alternates between reading and writing actions while receiving speech input in a continuous stream. The existing approach implicitly assumes that the time spent on generating text and processing streaming speech occurs sequentially. As a result, this can lead to unreliable accumulation of delay calculations.}
    \label{fig:fig3}
\end{figure}

In the simultaneous speech translation task, the system starts generating a hypothesis with only partial input. It alternates between reading a new source speech segment or writing a new target text segment. Assuming $\mathbf{X}_{1} = [x_1, ..., x_j]$, $j < |\mathbf{X}|$ has been read when generating $y_i$, where $x_j$ represents a raw audio segment of duration $T_j$. The Computation Aware (CA) and Computation Unaware (CU) delay of a token $y_i$ are defined as:
\begin{equation}
d_i = \begin{cases}
\sum_{k=1}^j T_k, & \text{CU} \\
\sum_{k=1}^j T_k + C_i, & \text{CA} \\
\end{cases}\end{equation}
where $C_i$ is the elapsed time when generating the $i$-th token, as recorded by SimulEval \cite{ma-etal-2020-simuleval} after generating this token.


The latency metrics are calculated using a normalization function, which takes the sequence of delays, \(\mathbf{D} = [d_1, \ldots, d_{|\mathbf{Y}|}]\), from SimulEval, along with a corresponding set of oracle delays, \(\mathbf{D}^* = [d^*_1, \ldots, d^*_{|\mathbf{Y}|}]\). This process can be formalized as follows:

\begin{equation}
Latency = \frac{1}{\tau'(|\mathbf{X}|)} \sum_{i=1}^{\tau'(|\mathbf{X}|)} (d_i - d_i^*), \label{eq:al}
\end{equation}

where $\tau'(|\mathbf{X}|) = \text{min}\{i|d_i = \sum_{j=1}^{|\mathbf{X}|}T_j\}$ and $d_i^*$ represents the oracle delays. \citet{ma-etal-2020-simuleval} suggests using $d_i^* = (i - 1) \cdot \frac{\sum_{j=1}^{|\mathbf{X}|} T_j}{|\mathbf{Y}^*|}$ to mitigate potential under-generation in Average Lagging (AL) measures. To correct the bias towards over-generation, \citet{papi-etal-2022-generation} recommends substituting $|\mathbf{Y}^*|$ with the larger value between $|\mathbf{Y}^*|$ and $|\mathbf{Y}|$ when computing oracle delays.

\section{The Problem of current CA}
\label{sec:section3}

\begin{figure*}[htbp]
    \centering
    \includegraphics[width=1.0\textwidth]{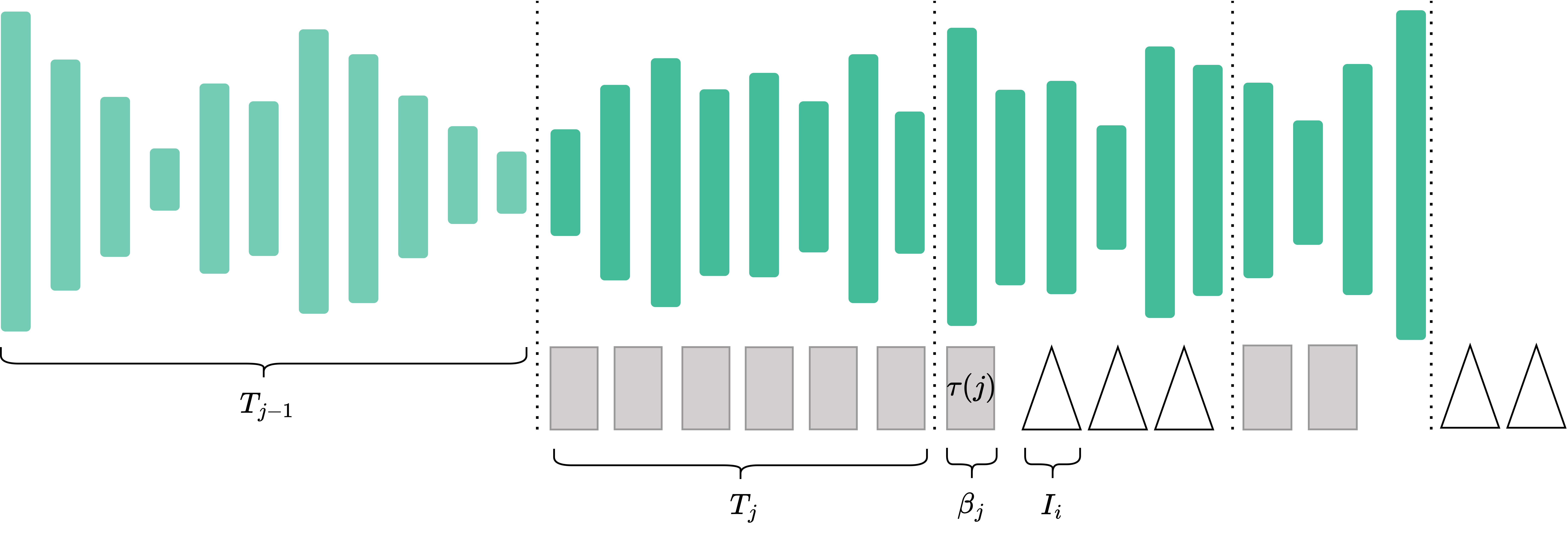}
    \caption{Previous generation time exceeding the current segment $x_j$ introduces additional delay for generating tokens in the current frame. For the given example, if generating each token takes 1 second, after reading 1 second of speech, the system takes 2 seconds to generate 'ein Mississippi', resulting in $\beta_j$ being 1 second. The delay for $I_i$ would be the sum of the previous speech duration, $\beta_j$, and $I_i$. If there is no buffer, later tokens are not affected.}
    \label{fig:main}
\end{figure*}


In practice, the system receives new speech input while generating text translations. However, as shown in Figure \ref{fig:fig3}, the CA delay $\sum_{k=1}^j T_k + C_i$ treats a parallel reading and writing process sequentially, assuming the system alternates between reading and writing actions. As a result, the CA delay accumulates the computation cost at each step and deviates from the real-world computation-aware delay, which in turn makes the latency measurement in Equation \ref{eq:al} unreliable.

To illustrate this more concretely, let us consider an example. Suppose we have a source sequence consisting of three segments, each representing the phrase "one Mississippi," which takes approximately one second to say. Thus, $ \mathbf{X} = [x_1, x_2, x_3] $ consists of three speech segments, each lasting one second ($ T_1 = T_2 = T_3 = 1 $ s). The system decides to generate output after processing $ x_1 $, and it takes 0.5 second to generate each token. For an English-to-German translation, the system generates the translation "ein" at the timestamp of 1.5 seconds and "Mississippi" at 2 seconds.

Intuitively, after processing all the source segments, the total delay for generating the subsequent tokens $y_5$ and $y_6$ should be $ \sum_{k=1}^3 T_k + 0.5 = 3.5 $ seconds and 4 seconds, respectively. However, with the current method, the computation timestamps $C_i$ for the third segment are calculated with accumulated computation costs at previous steps. The generation times $C_i$ for the fifth and sixth tokens would be 2.5 seconds and 3 seconds, resulting in delays of 5.5 seconds ($3 + 2.5$) and 6 seconds ($3 + 3$) for the two tokens.
\section{Method}
\label{sec:section4}

We propose a revised way of calculating CA delay to fix this problem.
We define the \emph{inference time} \( I_i \) as the elapsed time since processing the previous source segment $x_{j-1}$. Formally, this is expressed as:

\begin{equation}
    I_i = C_i - C_{\tau(j)},
\end{equation}

where \( C_i \) is the computation timestamp when generating token \( y_i \), and \( C_{\tau(j)} \) is the computation timestamp associated with the reference token \( y_{\tau(j)} \). The index \( \tau(j) \) represents the last token generated before processing the current source segment \( x_j \), defined as:

\begin{equation}
    \tau(j) = \max\left\{ i \mid d_i \leq \sum_{k=1}^{j-1} T_k \right\}.
\end{equation}

Here, \( d_i \) is the theoretical computation unaware delay at token \( y_i \), and \( \sum_{k=1}^{j-1} T_k \) is the cumulative duration of the source segments up to \( x_{j-1} \).

To represent the accumulated delay effect caused by discrepancies between computation time and source segment durations, we introduce the a buffer \( \beta_j \) corresponding to the source segment \( x_j \):

\begin{equation}
    \beta_j = \max\left\{ 0, \beta_{j-1} + I_{\tau(j)} - T_j \right\},
\end{equation}

where \( \beta_{j-1} \) is the buffer from the previous source segment \( x_{j-1} \) that affects the accumulation status while processing segment \( x_{j} \), and \( T_j \) is the duration of the current source segment \( x_j \). The buffer accumulates any excess inference time that is not covered by the source segment durations.

Finally, the computation-aware delay \( d_i \) for token \( y_i \) can be defined as:

\begin{equation}
    d_i = \beta_j + I_i + \sum_{k=1}^j T_k,
\end{equation}
which is calculated by combining the accumulation status buffer \(\beta\), the inference time \( I \), and the cumulative source segment duration.
\section{Experiment}
\label{sec:section5}

To assess the effectiveness of our proposed modification in aligning with the real-world performance of SimulST systems, we simulated a streaming speech translation task using a 25-second input. The average completion time was 27,020 ms\footnote{We built the service based on Gradio and measured the finish time by detecting the EOS token. The experiment was run five times, employing a wait-k stride-n policy and a Conformer-based encoder-decoder offline ST model.}. The SimulEval recorded last token's CA* delay calculated by our proposed method was 26,311 ms, \textbf{resulting in a difference within 2\%}. In contrast, the last token CA delay without our modification was 39,602 ms, \textbf{resulting in a difference of 46.6\%}.

\begin{table}[ht]
\centering
\begin{tabular}{l|c c c c}
\hline
\textsc{Frame}  & 2    & 6    & 10   & 12   \\ \hline
\textsc{CU}     & 1256 & 1668 & 2144 & 2582 \\ 
\textsc{CA}     & 1897 & 2317 & 2874 & 3391 \\ 
\textsc{CA*}    & 1617 & 2008 & 2551 & 3063 \\ \hline
\end{tabular}
\caption{AL in ms, we use AlignAtt as SimulST system's policy and set the frame to 2, 6, 10, 12 to represent different latency setting.}
\label{table:AL}
\end{table}

\begin{table}[ht]
\centering
\begin{tabular}{l|c c c c}
\hline
\textsc{Frame}  & 2    & 6    & 10   & 12   \\ \hline
\textsc{CU}     & 951 & 1459 & 1983 & 2448 \\ 
\textsc{CA}     & 1675 & 2167 & 2763 & 3306 \\ 
\textsc{CA*}    & 1346 & 1823 & 2411 & 2950 \\ \hline
\end{tabular}
\caption{LAAL in ms, we use AlignAtt as SimulST system's policy and set the frame to 2, 6, 10, 12 to represent different latency setting.}
\label{table:LAAL}
\end{table}

\paragraph{Effects on Pre-Segmented Speech}
Due to the incorrect calculation of computation-aware delay, the computation-aware latency deviates from the real latency, and this deviation increases as speech length increases, as shown in Figure~\ref{fig:intro}. However, even on pre-segmented short speeches, the deviation caused by CA delay is still large.

We use the MuST-C v1.0 En-De tst-COMMON~\cite{di-gangi-etal-2019-must} to examine such effects. We utilize a Conformer-based encoder-decoder offline ST model combined with the AlignAtt policy~\cite{papi-etal-2023-attention}, which relies on cross-attention to determine whether to emit translated words or wait for additional information.

We identify severe CA computation issues with input lengths of 25 seconds. As illustrated in Tables ~\ref{table:AL} and \ref{table:LAAL}, even for pre-segmented speech averaging 5 seconds in length, the discrepancies in both Average Lagging (AL) and Length-Adaptive Average Lagging (LAAL) are greater than 300 ms.

\section{Conclusion}
In this work, we investigated the shortcomings of current latency evaluation metrics for SimulST systems, focusing on the discrepancies caused by computation-aware delay miscalculations. We demonstrated that these inaccuracies lead to unrealistic latency estimates,  not only in long streaming speech translation but also in pre-segmented speech. Our proposed modification addresses these misconceptions and aligns the latency calculations more accurately with the SimulST system's real behavior, leading to a more reliable evaluation for both SimulST and StreamingST systems.


\section*{Limitations}

This work introduces CA* to improve the accuracy of latency evaluation for SimulST systems by addressing computation-aware delay. However, other limitations remain. Notably, oracle delay approximations may inaccurate, especially for unsegmented streaming speech, which includes long pauses and varied segments. Future work should refine these approximations to reduce errors.

Furthermore, this study focused on AL and LAAL. Although CA* can be generalized to other metrics, additional evaluation is required to confirm its effectiveness across various latency measures.

We also used ChatGPT for grammar revisions.

\bibliography{anthology,custom}

\appendix

\end{document}